# Single Neuron Memories and the Network's Proximity Matrix

Subhash Kak
Oklahoma State University, Stillwater

**Abstract:** This paper extends the treatment of single-neuron memories obtained by the use of the B-matrix approach. The spreading of activity within the network is determined by the network's proximity matrix which represents the separations amongst the neurons through the neural pathways.

## Introduction

It is reasonable to assume that in a network of densely connected neurons, activity originates in some area and spreads to others. Neurons belonging to different physiological structures are associated with different functions and their organization is characterized by category and hierarchy [1]. As to neuron activity, after firing, they enter a phase of refractoriness. Conversely, models of neural network memory have feedback connections trained by Hebbian learning [2],[3] without consideration to physical separation amongst the neurons and the delay caused by the propagation of spike activity. The B-matrix approach to neural network function [4] accounts for spreading of activity from one region to others based on adjacency of neurons. The B-matrix approach emerged out of research seeking to find ways to index information in neural networks [5]-[11].

Feedforward networks [12]-[14] are naturally associated with spreading activity because the direction of activity is specified *a priori*. The B-matrix model extends the notion of spreading of activity to feedback networks. The order in which the activity will spread depends on the propagation time for the spiking activity to reach neighboring neurons, which is reflected in the proximity values associated with the network.

We will show in this paper that since the proximity matrix defines different activity orders related to different starting neurons, it makes it possible for the neural network to associate different memories with different neurons.

## The Proximity Matrix

The physical organization of any neural network makes the separations amongst the specific neurons to have many different values. The activity would have a tendency to flow in different directions for different starting points. It is also important to recognize that the neurons separations need not satisfy Cartesian constraints since neural pathways may be coiled up or twisted in a three-dimensional geometry. The proximity matrix $P$ represents the distances between all pairs of neurons.

We will consider two simple examples to illustrate proximity distances. The *(ij)* values in this matrix represent the distance from neuron $i$ to neuron $j$.

**Example 1.** Consider four neurons associated with the following matrix:

$$P = \begin{bmatrix} 0 & 1 & 2 & 4 \\ 1 & 0 & 1.5 & 2 \\ 2 & 1.5 & 0 & 3 \\ 4 & 2 & 3 & 0 \end{bmatrix} \qquad (1)$$

Although, neurons #1 and #2 are 1 unit apart and neurons #1 and #3 are 2 units apart, neurons #2 and #3 are 1.5 units apart. This is illustrated in Figure 1 below.

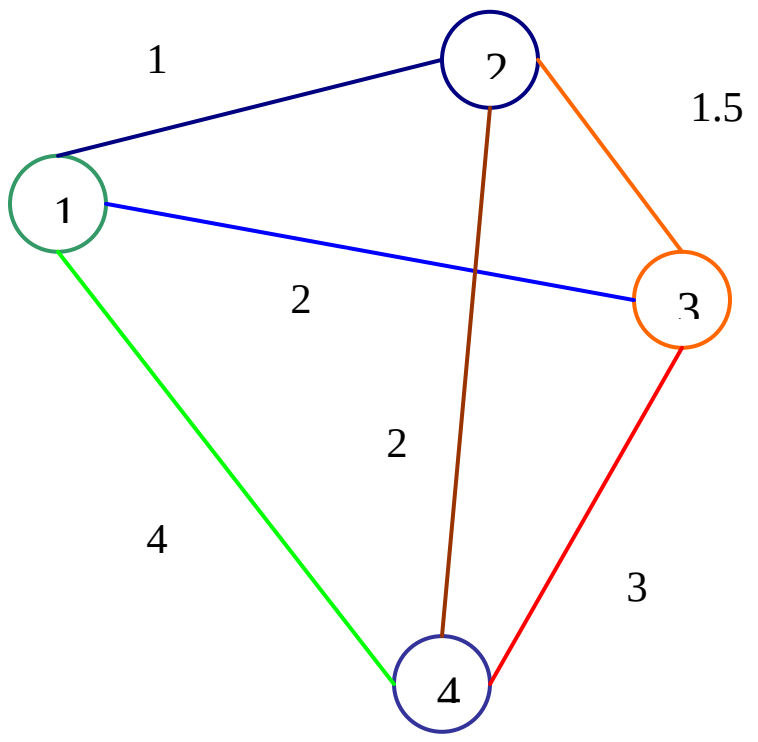

**Figure 1:** Separations between the four neurons of Example 1
(separations are not shown to scale)

The order of activity spread for the four neurons is as follows:

*Neuron 1*: 1 2 3 4
*Neuron 2*: 2 1 3 4
*Neuron 3*: 3 2 1 4
*Neuron 4*: 4 2 3 1

We notice that each neuron is characterized by a unique order. None of the four orders is a cyclic permutation of the other.

**Example 2.** Consider, next, the proximity matrix $P$ for five neurons:

$$P = \begin{bmatrix} 0 & 1 & 2.5 & 4 & 7 \\ 1 & 0 & 2 & 4.5 & 3 \\ 2.5 & 2 & 0 & 1 & 6 \\ 4 & 4.5 & 1 & 0 & 5 \\ 7 & 3 & 6 & 5 & 0 \end{bmatrix} \qquad (2)$$

Thus neurons 1 and 2 are one unit apart from each other, neurons 1 and 3 are 2.5 units apart, neurons 1 and 4 are 4 units apart, and neurons 1 and 5 are 7 units apart.

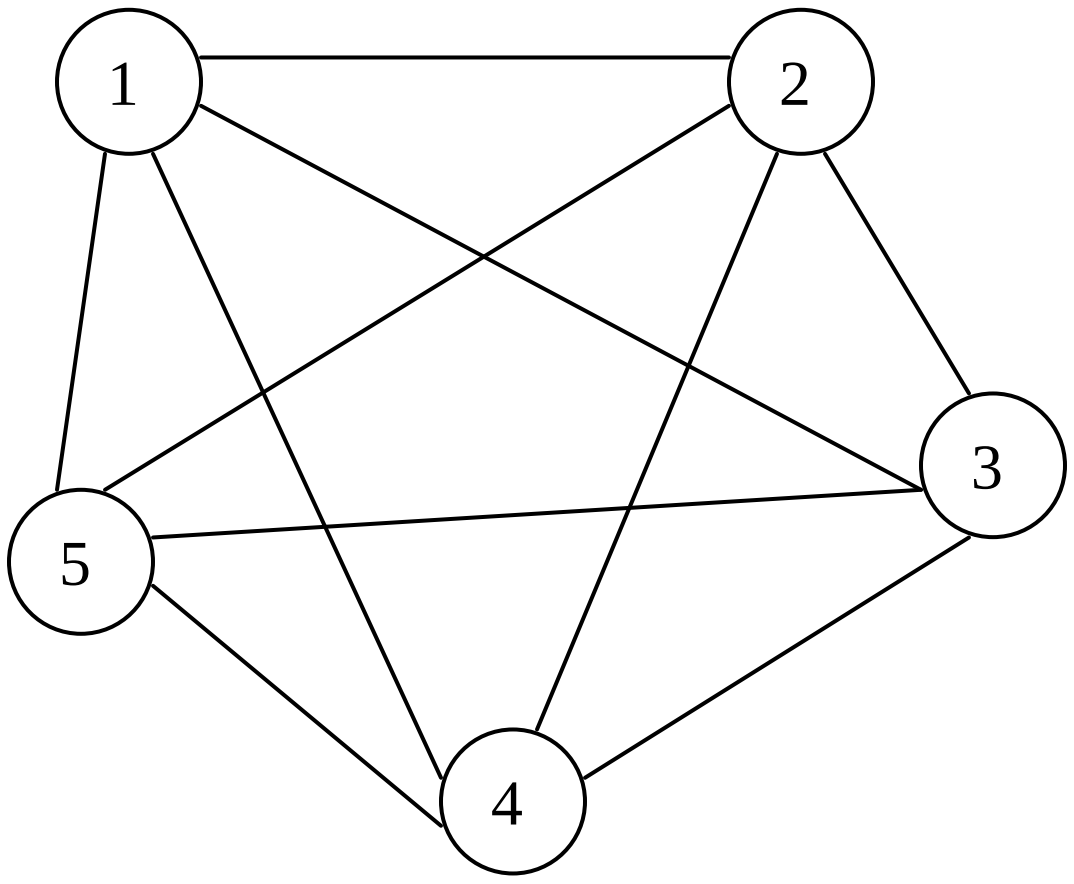

**Figure 2.** Network of five neurons

This implies that starting with neuron 1, the flow of activity will be 1 2 3 4 5; starting with neuron 2, it will be 2 1 3 5 4; starting with neuron 3, it will be 3 4 2 1 5; starting with neuron 4, it will be 4 3 1 2 5; and starting with neuron 5, it will be 5 2 4 3 1.

The order of activity flow varies considerably for the five neurons. Once again none of these orders is a cyclic permutation of the other.

## Hebbian learning and the B-matrix

The neural network is trained by Hebbian learning, in which the connectivity (synaptic strength) of neurons that fire together strengthens and that of those that don't gets weakened. The interconnection matrix $T = \sum x^{(i)} x^{(i)t}$, where the memories are column vectors, $x^i$, and the diagonal terms are taken to be zero.

A memory is stored if

$$x^i = \text{sgn}(Tx^i) \quad (3)$$

where the *sgn* function is 1 if the input is equal or greater than zero and -1, if the input is less than 0. We know from experiments that the memory capacity of such a network is about 0.15 times the number of neurons. In addition to the stored desired memories, their complements will be stored (not all because of the asymmetry of the sgn operation), as also some spurious memories [3],[10],[11].

In the generator model [4],[5], the memories are recalled by the use of the lower triangular matrix *B*, where $T = B + B^t$. Effectively, the activity starts from one single neuron and then spreads to additional neurons as determined by *B* (Figure 3). Note that with each update, the fragment enlarges by one neuron and it is fed back into the circuit. The spreading function where the information spreads from neuron 1 to neuron 2 and so on has an implicit assumption regarding the geometrical (proximity) relationship amongst the neurons.

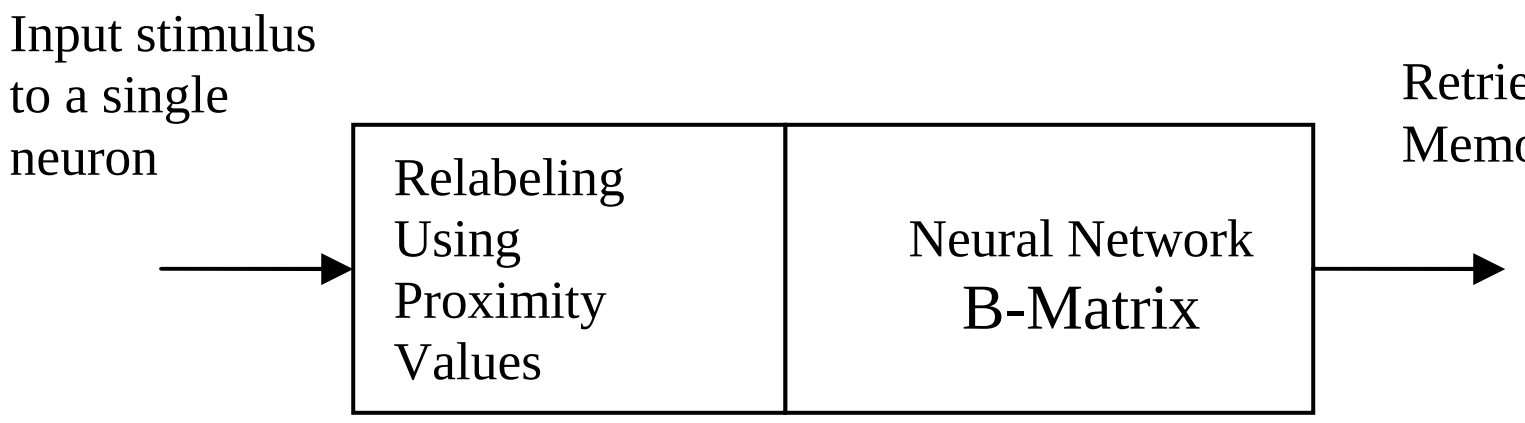

**Figure 3.** Use of proximity values to determine the neural memory

For the spreading of activity from any specific neurons, one must first compute the proximity values to the other neurons to determine how the activity will spread. This was shown for Examples 1 and 2 above.

Starting with the fragment $f^i$, the updating proceeds as:

$$f^i\ (new) = \text{sgn}\ (Bf^i\ (old)\ ) \quad (4)$$

where the original fragment values are *left unchanged* on the neurons and the updating proceeds *one step at a time*. Thus the activity will expand from one to two neurons and then on to three neurons, and so on.

This will be illustrated by means of the following example.

**Example 3.** Consider Hebbian learning applied to the following three memories, $x^i$, $i=1,2,3$, where each memory is a column vector that is shown below as its transpose:

$$x^{1t} = 1 \quad 1 \quad 1 \quad 1 \quad 1$$
$$x^{2t} = 1 \quad -1 \quad -1 \quad -1 \quad 1$$
$$x^{3t} = 1 \quad 1 \quad -1 \quad -1 \quad -1 \qquad (5)$$

We assume that the proximity matrix associated with the five neurons is as given in the previous Example 2 with the corresponding equation (2).

The interconnection weight matrix is the *T*-matrix in equation (6). It may be easily checked that all the three memories are indeed stored for this example.

$$T = \begin{bmatrix} 0 & 1 & -1 & -1 & 1 \\ 1 & 0 & 1 & 1 & -1 \\ -1 & 1 & 0 & 3 & 1 \\ -1 & 1 & 3 & 0 & 1 \\ 1 & -1 & 1 & 1 & 0 \end{bmatrix} \qquad (6)$$

As explained above, the dynamics of the spreading function will be governed by the triangular matrix *B*:

$$B = \begin{bmatrix} 0 & 0 & 0 & 0 & 0 \\ 1 & 0 & 0 & 0 & 0 \\ -1 & 1 & 0 & 0 & 0 \\ -1 & 1 & 3 & 0 & 0 \\ 1 & -1 & 1 & 1 & 0 \end{bmatrix} \qquad (7)$$

Here it is assumed that the first value (on the specified neuron) spreads to the next neuron, and then the two together, spread to the third, and so on. The spreading is governed by the *B* matrix.

**Activity starting at Neuron 1**

Let bit "1" be clamped on neuron number 1, and, therefore, in the updating below, this value will not change. The updating order is 1 2 3 4 5 which means that we do not have to change the labels of the five neurons.

$$f^{1}(new) = \operatorname{sgn} \begin{bmatrix} 0 & 0 & 0 & 0 & 0 \\ 1 & 0 & 0 & 0 & 0 \\ -1 & 1 & 0 & 0 & 0 \\ -1 & 1 & 3 & 0 & 0 \\ 1 & -1 & 1 & 1 & 0 \end{bmatrix} \begin{bmatrix} 1 \\ 0 \\ 0 \\ 0 \\ 0 \end{bmatrix} = \begin{bmatrix} 1 \\ 1 \\ 0 \\ 0 \\ 0 \end{bmatrix} \Longrightarrow \begin{bmatrix} 1 \\ 1 \\ 1 \\ 1 \\ 1 \end{bmatrix} \qquad (8)$$

At the second pass, $f^1$ *(new)= [1 1 1 0 0]$^t$*. The operation in (8) proceeded on a neuron-by-neuron basis. The fragment increased from "1" to "1 1" and it was fed back into the circuit and increased to "1 1 1" and so on until Memory #1 is retrieved.

**Activity starting at Neuron 2**

The updating order is 2 1 3 5 4. The T-matrix with respect to this updating of the neuron orders is:

$$T^{(2)} = \begin{bmatrix} 0 & 1 & 1 & -1 & 1 \\ 1 & 0 & -1 & 1 & -1 \\ 1 & -1 & 0 & 1 & 3 \\ -1 & 1 & 1 & 0 & 1 \\ 1 & -1 & 3 & 1 & 0 \end{bmatrix} \qquad (9)$$

This is the same matrix as that of equation (6) excepting that neurons 1 2 3 4 5 have been permuted to 2 1 3 5 4.

The corresponding triangular matrix $B^{(2)}$:

$$B^{(2)} = \begin{bmatrix} 0 & 0 & 0 & 0 & 0 \\ 1 & 0 & 0 & 0 & 0 \\ 1 & -1 & 0 & 0 & 0 \\ -1 & 1 & 1 & 0 & 0 \\ 1 & -1 & 3 & 1 & 0 \end{bmatrix} \qquad (10)$$

$$f^2(new) = \mathrm{sgn} \begin{bmatrix} 0 & 0 & 0 & 0 & 0 \\ 1 & 0 & 0 & 0 & 0 \\ 1 & -1 & 0 & 0 & 0 \\ -1 & 1 & 1 & 0 & 0 \\ 1 & -1 & 3 & 1 & 0 \end{bmatrix} \begin{bmatrix} 1 \\ 0 \\ 0 \\ 0 \\ 0 \end{bmatrix} \Longrightarrow \begin{bmatrix} 1 \\ 1 \\ 1 \\ 1 \\ 1 \end{bmatrix} \qquad (11)$$

Since all the neuron values are identical, relabeling of the neurons is not required and we clearly have obtained Memory #1.

**Activity starting at Neuron 3**

The updating order now is 3 4 2 1 5, with the corresponding $T^{(3)}$ interconnection weight matrix:

$$T^{(3)} = \begin{bmatrix} 0 & 3 & 1 & -1 & 1 \\ 3 & 0 & 1 & -1 & 1 \\ 1 & 1 & 0 & 1 & -1 \\ -1 & -1 & 1 & 0 & 1 \\ 1 & 1 & -1 & 1 & 0 \end{bmatrix} \quad (12)$$

The corresponding triangular matrix $B^{(3)}$:

$$B^{(3)} = \begin{bmatrix} 0 & 0 & 0 & 0 & 0 \\ 3 & 0 & 0 & 0 & 0 \\ 1 & 1 & 0 & 0 & 0 \\ -1 & -1 & 1 & 0 & 0 \\ 1 & 1 & -1 & 1 & 0 \end{bmatrix} \quad (13)$$

$$f^{3}(new) = \operatorname{sgn} \begin{bmatrix} 0 & 0 & 0 & 0 & 0 \\ 3 & 0 & 0 & 0 & 0 \\ 1 & 1 & 0 & 0 & 0 \\ -1 & -1 & 1 & 0 & 0 \\ 1 & 1 & -1 & 1 & 0 \end{bmatrix} \begin{bmatrix} -1 \\ 0 \\ 0 \\ 0 \\ 0 \end{bmatrix} \Rightarrow \begin{bmatrix} -1 \\ -1 \\ -1 \\ 1 \\ 1 \end{bmatrix} \quad (14)$$

In other words, for the news labeled 3 4 2 1 5, the bits obtained are -1 -1 -1 1 1, which, when mapped to the normative order implies 1 -1 -1 -1 1, which, is Memory #2.

**Activity starting at Neuron 4**

The updating order now is 4 3 1 2 5, with the corresponding $T^{(4)}$ interconnection weight matrix:

$$T^{(4)} = \begin{bmatrix} 0 & 3 & 1 & -1 & 1 \\ 3 & 0 & 1 & -1 & 1 \\ 1 & 1 & 0 & 1 & -1 \\ -1 & -1 & 1 & 0 & 1 \\ 1 & 1 & -1 & 1 & 0 \end{bmatrix} \quad (15)$$

which is identical to the case of activity starting with Neuron 3. Therefore, this case also we will end up in Memory #2.

**Activity starting at Neuron 5**

The updating order now is 5 2 4 3 1, with the corresponding $T^{(5)}$ interconnection weight matrix:

$$T^{(5)} = \begin{bmatrix} 0 & -1 & 1 & 1 & 1 \\ -1 & 0 & 1 & 1 & 1 \\ 1 & 1 & 0 & 3 & -1 \\ 1 & 1 & 3 & 0 & -1 \\ 1 & 1 & -1 & -1 & 0 \end{bmatrix} \tag{16}$$

The corresponding triangular matrix $B^{(5)}$:

$$B^{(5)} = \begin{bmatrix} 0 & 0 & 0 & 0 & 0 \\ -1 & 0 & 0 & 0 & 0 \\ 1 & 1 & 0 & 0 & 0 \\ 1 & 1 & 3 & 0 & 0 \\ 1 & 1 & -1 & -1 & 0 \end{bmatrix} \tag{17}$$

$$f^{5}(new) = \operatorname{sgn} \begin{bmatrix} 0 & 0 & 0 & 0 & 0 \\ -1 & 0 & 0 & 0 & 0 \\ 1 & 1 & 0 & 0 & 0 \\ 1 & 1 & 3 & 0 & 0 \\ 1 & 1 & -1 & -1 & 0 \end{bmatrix} \begin{bmatrix} 1 \\ 0 \\ 0 \\ 0 \\ 0 \end{bmatrix} \Longrightarrow \begin{bmatrix} 1 \\ -1 \\ 1 \\ 1 \\ -1 \end{bmatrix} \tag{18}$$

In other words, for the neurons labeled 5 2 4 3 1, the bits obtained are 1 -1 1 1 -1, which, when mapped to the normative order implies -1 -1 1 1 1. This is the complement of Memory #3. If we had started with -1, we would have ended up with -1 1 1 1 -1, which is the complement of Memory #2.

**Discussion**

We see that upon consideration of proximity matrix associated with the neurons, the memories are obtained by starting with activity at specific neurons. We believe that such an activity appears more reflective of what happens in physical systems than one where all the neurons are taken to resonate by synchronized action or even in asynchronous updating that proceeds in all directions.

It should be stressed that network models of memory are limited in explaining various facets of human cognitive and computational capacity [15]-[19]. Our paper does not address these limitations any more than other network models do. But it builds a bridge between models of feedback neural networks where memory resides in resonant activity and networks in which the spread of activity is determined by physical propagation constraints.